\pdfoutput=1
\documentclass[11pt]{article}

\usepackage[final]{acl}
\usepackage{subcaption}
\usepackage{times}
\usepackage{latexsym}
\usepackage{multirow}%
\usepackage{xcolor}%
\usepackage{color, colortbl}
\usepackage{booktabs}%

\definecolor{Gray}{gray}{0.9}

\usepackage[T1]{fontenc}
\usepackage[utf8]{inputenc}

\usepackage{microtype}

\usepackage{inconsolata}

\usepackage{graphicx}

\usepackage{fdsymbol}

\newcommand\nyutwo{$^\vardiamondsuit$}
\newcommand\mitone{$^\spadesuit$}
\newcommand\ucla{$^\varheartsuit$}

\title{MisinfoEval: \\ Generative AI in the Era of ``Alternative Facts''}

\author{Saadia Gabriel\ucla \space\space\space Liang Lyu\mitone \space \space\space James Siderius\nyutwo \\ \textbf{Marzyeh Ghassemi}\mitone \space \space\space \textbf{Jacob Andreas}\mitone \space \space\space \textbf{Asu Ozdaglar}\mitone \\ \ucla University of California, Los Angeles \\  \mitone Massachusetts Institute of Technology \\ \nyutwo Dartmouth College, Tuck School of Business  }

\begin{document}
\maketitle
\begin{abstract}
The spread of misinformation on social media platforms threatens democratic processes, contributes to massive economic losses, and endangers public health. Many efforts to address misinformation focus on a knowledge deficit model and propose interventions for improving users’ critical thinking through access to facts. Such efforts are often hampered by challenges with scalability, and by platform users’ personal biases.  The emergence of generative AI presents promising opportunities for countering misinformation at scale across ideological barriers. 

In this paper, we introduce a framework (MisinfoEval) for generating and comprehensively evaluating large language model (LLM) based misinformation interventions. We present (1) an experiment with a simulated social media environment to measure effectiveness of misinformation interventions, and (2) a second experiment with personalized explanations tailored to the demographics and beliefs of users with the goal of countering misinformation by appealing to their pre-existing values. Our findings confirm that LLM-based interventions are highly effective at correcting user behavior (improving overall user accuracy at reliability labeling by up to 41.72\%). Furthermore, we find that users favor more personalized interventions when making decisions about news reliability and users shown personalized interventions have significantly higher accuracy at identifying misinformation. 
\end{abstract}

\section{Introduction}

In the last decade, there has been growing concern about the proliferation of misinformation on social media platforms. For example, between 2006 and 2017, sensational “fake news” articles spread rapidly on Facebook, diffusing farther and faster than truthful or reputable content \cite{doi:10.1126/science.aap9559}. These sharing trends have been amplified by “filter bubble” algorithms that intentionally create ideological echo chambers, which reinforce existing viewpoints and further facilitate spread of misinformation \cite{10.1257/aer.20191777,acemoglu}. 

Many proposed interventions for combating misinformation focus on tagging unreliable content \cite{Clayton2020RealSF,Pennycook2019TheIT} or encouraging critical thinking by users \cite{LUTZKE2019101964,Pennycook2019ShiftingAT}. However, the two major bottlenecks in many such interventions are user bias and scalability. Conventional fact-checking interventions rely on the assumption that users are rational agents, who will agree upon a common “ground truth” once exposed to enough information. This assumption is often violated in the real world due to the fact users do not process new information neutrally, and are more critical of counter-partisan news and more accepting of pro-partisan news at face value \cite{Lord1979BiasedAA,Nickerson1998ConfirmationBA,Tappin2019BayesianOB}. Tagging unreliable or suspicious content also requires careful inspection, which is currently performed by professional fact-checking organizations such as Snopes. These organizations are constrained both in terms of their financial resources and qualified fact-checkers they can employ. Alternatively, platforms have explored a decentralized approach through crowdsourced verification by users. While this approach allows for scale, it is vulnerable to user misuse such as recent reports of disinformation and partisan bias spread through X's Community Notes \cite{communityx}.

Breakneck advances in large language models (LLMs) offer a promising avenue for large-scale fact checking, as they provide tools for fast information processing and can detect patterns associated with misleading content \cite{Chen2023CombatingMI}. Early evidence suggests that LLM-based explanations of veracity can significantly reduce social media users’ tendency to accept false claims \cite{hsu-etal-2023-explanation}. LLMs also present a potential path to understanding and countering user bias: recent work \cite{andreas-2022-language,10.1145/3514094.3534177,gabriel-etal-2022-misinfo} argues LLMs are capable of very simple forms of world and cognitive modeling. This opens up the possibility of tailored approaches to countering misinformation that target users across diverse backgrounds (e.g. varying education levels or ideological leanings). Our agenda is to develop a powerful, automated tool via LLMs for generating tailored misinformation interventions and a framework for testing the effectiveness of these interventions (MisinfoEval). To do this, we have two study phases focused on examining effects of AI interventions with and without personalization: 

\paragraph{Phase I:} We conduct an A/B testing experiment using a simulated social media environment that features both true and false content. 

\paragraph{Implications for Content Moderation:} First we expand upon existing research showing that explanation-based credibility indicators can mitigate spread of misinformation. In a large-scale experiment, we show \textbf{explanation interventions can effectively inform diverse users about unreliable content, improving user accuracy over label-only indicators by at least 34.2\% vs. 24.2\% for label-only indicators}. Explanations generated by GPT-4 \cite{Achiam2023GPT4TR} perform best at encouraging user flagging of misinformation (pre-intervention users correctly flag in 3.1\% of cases vs. 38.1\% post-intervention). This indicates the promise of LLM-based explanations for future intervention strategies, corroborating findings from recent and concurrent work \cite{gabriel-etal-2022-misinfo,hsu-etal-2023-explanation}. 

\paragraph{Implications for Healthcare:} The spread of medical misinformation poses a serious obstacle to healthcare providers if unchecked. Some notable recent examples include false claims spread online through websites like X (formerly Twitter), YouTube and WebMD of ``alternative cures" for Cancer \cite{SwireThompson2019PublicHA}. Such false claims could pose a significant health risk given that alternative medicines can more than double the risk of mortality in Cancer patients \cite{Johnson2017UseOA}. \textbf{We find that GPT-4 explanations could provide automated support in mitigating medical misinformation, leading to an user accuracy at predicting news reliability of 97.6\% (95\% CI: [96.0, 99.2]).}

\paragraph{Phase II:} Next, we explore how personalization of explanations can further improve their effectiveness. We measure how the degree of personalization affects user-reported helpfulness, and find that \textbf{explanations that are highly aligned with users’ attributes (e.g., education, political ideology, gender) are deemed more helpful by users than explanations without personalization (average helpfulness score of 2.98 vs. 2.71)}. We also find that \textbf{users shown personalized explanations have significantly higher accuracy at identifying misinformation than users shown misaligned explanations}.

Our results highlight the powerful persuasive capabilities of LLMs. Our simulated social media feed platform and surveys with over 4,000 diverse participants will be made publicly available to further research on how LLM outputs influence users, including potential disinformation risks. To the best of our knowledge, this is the first study to consider tailored LLM-based interventions based on specific attributes of social media users. We view it as a first step in this agenda, since general advances in foundation models will increase these tools’ capabilities for combating misinformation. We envisage better fine-tuning of explanations as additional information becomes available about users (without violating their privacy). 

\begin{figure*}
\centering
\begin{subfigure}{.75\textwidth}
  \centering
  \includegraphics[width=1\linewidth]{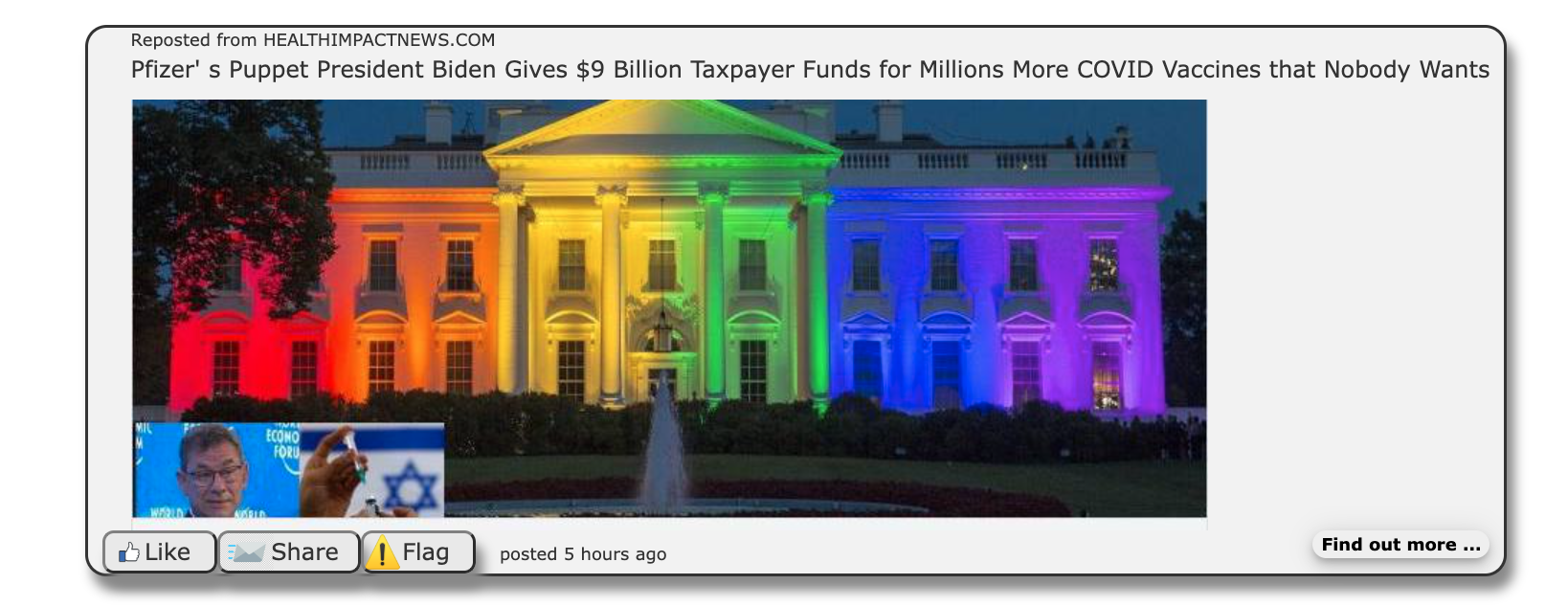}
  \label{fig:sub1}
\end{subfigure}%
\begin{subfigure}{.25\textwidth}
  \centering
  \includegraphics[width=1\linewidth]{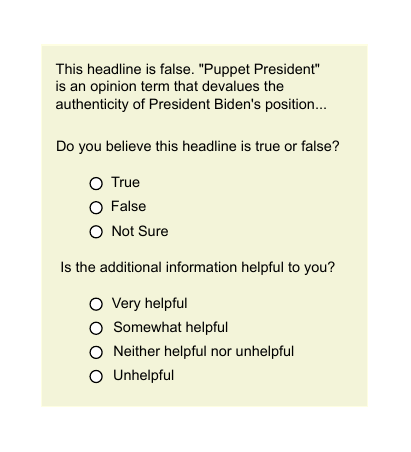}
  \label{fig:sub2}
\end{subfigure}
\caption{Examples of a post in the simulated newsfeed (left), and a pop-up intervention with a veracity label (right).}
\label{fig:fig2}
\end{figure*}

\section{Related Work}\label{background}

%In this section, we provide background definitions for misinformation or misleading content. We then place our work in the context of existing literature on mitigating misinformation since the advent of LLMs.
%\paragraph{Defining Misinformation.} By \textit{misinformation}, we refer to any content that is objectively false or misleading according to fact-checking sources (e.g. Snopes and Poynter). All reliability labels in this work were sourced from \citet{pennycook2021a}. In contrast to \textit{disinformation} (which is known by the author to be false), misinformation may consist of either intentionally and unintentionally false content. 
%\paragraph{Automated Mitigation of Misinformation.} 
Our work aligns with a growing body of literature, mostly from political science, that examines the effectiveness of mitigating misinformation through user-facing interventions. Assuming a known “ground truth” label determined by human or AI fact-checkers, this literature aims  to reduce user consumption and interaction with false content by designing effective ways to present this information or to otherwise nudge users to consider them.
\\
\\
Fact-checking labels that are specifically attributed to AI have been shown to be effective as interventions in reducing user consumption of misinformation \cite{social_media_use_trust_technology}, though earlier studies found they are often less effective than labels attributed to other sources such as professional fact-checkers \cite{10.1145/3292522.3326012,10.1145/3313831.3376213,10.1093/jcmc/zmab013,effects_of_fact_checking_social_media_vaccine}. There is evidence that explaining the mechanics behind how the fact-checking label is generated improves their effectiveness \cite{Epstein2021DoEI}. Concurrently with our work, it has been found that GPT-based explanations of content veracity can significantly reduce social media users’ reported tendency to accept false claims \cite{hsu-etal-2023-explanation}, though they can be equally effective when used with malicious intent to generate deceptive explanations \cite{Danry2022DeceptiveAS}. There have also been some early works that explore the use of personalization in AI fact-checking systems, such as \citet{inproceedings}, which examines the effects of a personalized AI prediction tool based on the user’s own assessments; and \citet{10.1145/3610080}, with a focus on toxicity in personalized content moderation tools. Our work departs from these as we consider the generation of arguments and justifications given a label, rather than predicting the veracity.

\section{Social Media Platform Experiment and Study Design}\label{social_media_study}

We recruit human participants to interact with a simulated news feed interface that mimics real-world social media platforms such as Facebook and X. The news feed consists of news headlines (claims), and an intervention button (“Find out more”) that the user may voluntarily click on. Upon doing so, they may see an intervention with a veracity label of the headline (true or false) and an explanation of the label (see Figure \ref{table:tab1}). Users may react to the news item as they normally would on social media, and provide feedback on the intervention. By varying the types of interventions presented to users and comparing their subsequent behavior, we can analyze the impacts of these interventions. Details of the interface are given in §3.1. Phase I (§4.1) compares the effectiveness of five non-personalized explanations, while Phase II (§4.2) directly compares GPT-4 generated explanations with and without personalization. In both phases, the experimental interface consists of four components: (1) a consent form; (2) task instructions; (3) a questionnaire on user demographics and opinions; (4) a simulated newsfeed. 
\begin{center}
\begin{table*}[t]
\begin{tabular}{p{2cm} | p{7cm} | p{6cm} } 
 \toprule
\small Intervention Type & \small Description & \small Example  \\ [0.5ex] 
 \bottomrule \toprule
\small Label Only & \small A simple ground-truth label indicator   & \small \textbf{\textit{This claim is true/false.}}  \\ 
 \hline
 \small Methodology (AI) & \small Following from \citet{Epstein2021DoEI}, we show users a generic explanation which states AI to be the source of the claim veracity label. & \small \textbf{\textit{This claim was verified/refuted by an AI model trained on a large-scale corpus of web data.}} \\
 \hline
 \small Methodology (Human) & \small Same as above, except the source is stated to be fact-checkers. & \small \textbf{\textit{This claim was verified/refuted by non-partisan fact-checkers.}}\\
 \hline
 \small Reaction Frame Explanation & \small Following from \citet{gabriel-etal-2022-misinfo}, we show users a templated explanation constructed using GPT-2 \cite{Radford2019LanguageMA} predictions for the intent of the claim author as perceived by the reader and potential actions a reader may take in response to a claim. & \small \textbf{\textit{This claim is true/false. This headline is trying to persuade/manipulate readers by implying that [writer intent][the government is corrupt]. It is compelling readers to [reader action][want to find out more].}}\\
 \hline
 \small Zero-shot GPT-4 Explanation & \small We instruct GPT-4 to generate an explanation for why a claim is false/true without user-specific information. The prompt given to GPT-4 is “\textbf{\textit{Write a short explanation for why the headline ‘[claim]’ is ‘[label].’ Do not mention that you are AI. The explanation must be less than 100 words.}}”  & \small \textbf{\textit{The headline is false because Dr. Anthony Fauci has not been arrested by any Special Forces. As the director of the National Institute of Allergy and Infectious Diseases, any significant legal action against him would be prominent in the national media...}} \\ \hline
 \small Personalized GPT-4 Explanation & \small We do the same as above, but with the following prompt that incorporates inferred user attributes \textbf{\textit{“Write a short explanation for why the headline ‘[claim]}}[Special Forces Arrest Deep State Dr. Anthony Fauci]\textbf{\textit{’ is ‘[label]}}[false]\textbf{\textit{’ that will appeal to an [education]}}[uneducated]\textbf{\textit{, [gender]}}[male]\textbf{\textit{, [race]}}[white]\textbf{\textit{, [age]}}[18-29] \textbf{\textit{year old reader with [politics]}}[conservative] \textbf{\textit{political beliefs. Do not mention that you are AI. Do not mention the type of reader. The explanation must be less than 100 words.”}}   & \small \textbf{\textit{This headline is false because there's no evidence. Terms like 'Deep State' often get thrown around without proof. Remember, just because someone disagrees with Dr. Fauci's advice, doesn't mean he's part of a conspiracy. Don't believe everything you read on the internet, mate. Always consider the source.}}\\[1ex] 
 \bottomrule
\end{tabular}
\caption{Types of intervention methods used in this experiment. }
\label{table:tab1}
\end{table*}
\end{center}

\subsection{MisinfoEval Environment}

Each participant receives 5 news items, randomly sampled from a dataset of 460 news headlines collected by \cite{pennycook2021a}. Our experiment uses 188 true articles and 185 false articles. Each news item consists of the \textit{headline} (which we also call a \textit{claim}), the accompanying image and the source of the news article.\footnote{The dataset also includes misleading headlines, but we omit these to have a binary true/false label.} 

\paragraph{User Interface Design.} Users can interact with the posts by liking, sharing or flagging them (Figure 2, left). Each user is instructed to perform at least one of these interactions for at least three out of five news items. Users also have the option to click on a “Find out more” button, which displays a pop-up that we call an \textit{intervention}. Except for the control setting, the intervention consists of two pieces of information: a \textit{label} indicating whether the claim is true or false and an \textit{explanation} either supporting or refuting the claim based on the label. In the pop-up, users can rate the perceived helpfulness of this information on a 4-point Likert scale (very helpful, somewhat helpful, somewhat unhelpful, very unhelpful). They can also indicate whether they believe the claim is true, false or are uncertain.\footnote{The exception is the control setting in Phase I, where the pop-up only asks users to evaluate the reliability of the claim. } 

\paragraph{MisinfoEval Intervention Types.}  In Phase I, we consider 5 types of previously proposed interventions for misinformation mitigation. Each participant is randomly assigned to one of the five types of interventions with equal probability and only shown that intervention. In Phase II, we introduce a sixth intervention type: personalized GPT-4 interventions. Table \ref{table:tab1} lists the six types of interventions, with examples based on the false claim "\textit{Special Forces Arrest Deep State Dr. Anthony Fauci}". 

In Phase II, we compare non-personalized and personalized GPT-4 interventions (the last two rows of Table \ref{table:tab1}). Personalized interventions tailor to a specific demographic group based on a set of attributes (gender, race, age, education level, political affiliation). Details on participant selection and attribute values used for the personalization study can be found in A.5. 

\begin{center}
\begin{table*}[t]
\begin{tabular}{p{1.7cm}|c|c|c|c|c|c|c|p{1.7cm}} 
 \toprule
 \small
 Intervention  & \multicolumn{3}{|p{3cm}|}{\small Accuracy (\% Correct)} & \multicolumn{2}{|p{1.8cm}|}{\small False Content Sharing (\%)} & \multicolumn{2}{|p{1.8cm}|}{\small False Content  Flagging (\%)} & \small Helpfulness (\% Helpful or Very Helpful)\\ \cmidrule{2-8}
  & \small Before & \small After & $\Delta$ & \small Before  & \small After & \small Before & \small After & \\ [0.5ex] 
 \bottomrule\toprule
\small Label Only & \small 55.00 $\pm$ 2.66 & \small 79.25 $\pm$ 2.33 & \small 24.24 & \small 5.12 & \small 4.18 & \small 4.37 & \small 21.38 & \small 85.43\\ 
 \hline
\small Reaction Frame & \small 55.07 $\pm$ 2.09 & \small 95.84 $\pm$ 1.00 & \small 40.77 & \small 4.62 & \small 0.00 & \small 3.83 & \small 1.00 & \small 94.31 \\
 \hline
 \small GPT-4 (non-personalized) & \small 52.16 $\pm$ 3.60 &  \small 93.88 $\pm$ 1.54 & \small 41.72 & \small 3.66  & \small 7.26 & \small 3.10 & \small 38.17 & \small 92.17\\ \hline
  \small Methodology Explanation (AI) & \small 51.87 $\pm$ 2.60 & \small 90.98 $\pm$ \small 1.35 & \small 39.11 & \small 9.59 & \small 16.95 & \small 2.82 & \small 28.95 & \small 88.66\\ \hline
 \small Methodology Explanation (Human) & \small 55.77 $\pm$ 2.04  & \small 89.97 $\pm$ 1.45 & \small 34.20 & \small 2.36 & \small 5.60 & \small 5.34 & \small 4.23 & \small 90.14\\
 \bottomrule
\end{tabular}
\caption{Accuracy at ground-truth label prediction, changes in interactions and perceived helpfulness results for all intervention types, both before interventions (left column) and after interventions (right column). Accuracy is shown with 95\% bootstrapped confidence intervals.  }
\label{table:tab2}
\end{table*}
\end{center}

\subsection{Participants}

In this section, we explain our methodology for user recruitment and qualification tasks we require users to undergo in order to ensure quality of results (e.g., filtering spammers).  

\subsubsection{Recruitment and Quality Control}

We use the Amazon Mechanical Turk\footnote{\url{https://www.mturk.com/}} crowdsourcing platform to recruit a diverse pool of 9,262 workers from the United States with at least a 98\% HIT\footnote{Human intelligence task} approval rating as potential study participants. To filter spamming workers, we ask them two “attention checks” questions that require them to write out the minimum number of posts they must interact with (3) and the number of posts in the newsfeed (5). Any workers who fail either of these attention checks are disqualified from participating in the rest of the study. We also disqualify workers who fail to follow the instructions by interacting with less than 3 posts. Lastly, we filtered workers who completed over 10 HITs but always predict the same reliability label for news articles (either true or false). 4,950 workers passed the qualification tests. Volunteered information about worker demographics is given in \ref{sec:crowddemos}. 

\subsubsection{Selection of User Attributes for Personalization}

Prior work \cite{Santurkar2023WhoseOD} has shown that identity groups based on the attributes we use for personalization (e.g. gender and political affiliation) have divergent beliefs around key social and political issues, which may affect their perception of news reliability. While many of these attributes are indirectly associated with beliefs, we also personalize based on political ideology, which is directly associated with beliefs. We leave exploration of how personalization influences specific beliefs (e.g. vaccine hesitancy) to future work.

\section{MisinfoEval Environment Results}

In \ref{sec:discuss}, we discuss best practices for evaluating effectiveness of misinformation interventions. We first compare non-personalized interventions in a simulated social media newsfeed in \ref{sec:non-personalized}. Then we assess the effectiveness of personalized interventions in \ref{sec:personalized}. 

\subsection{Quantifying Effectiveness of Explanations}
\label{sec:discuss}

Progress in development of misinformation interventions has been hindered by a lack of standardization in evaluation. In line with the framework posited by \citet{Guay2023HowTT}, we measure user interaction with \textit{both} true and false content. Our aim is to capture \textit{discernment} - the extent to which an user believes or intends to share false content relative to true content. We quantify effectiveness of interventions using two metrics:
\begin{itemize}
\item The objective effect of interventions on users’ accuracy in recognizing misinformation and factual content.
\item Users’ perception of the interventions’ helpfulness in performing misinformation detection.
\end{itemize}  

While we also provide results for user interactions with claims, we emphasize accuracy and helpfulness as effectiveness metric given that interactions like sharing are not clear indicators of a user's belief in a claim \cite{Pennycook2019ShiftingAT}. 

\subsection{Phase I: Non-personalized Explanations}
\label{sec:non-personalized}

We measure the effectiveness of all five (non-personalized) interventions in mitigating misinformation, by comparing users’ pre- and post-intervention behavior. We use the following metrics for the comparison: (1) accuracy of users’ veracity prediction for each headline; (2) interaction with false headlines, such as sharing and flagging; (3) user-reported helpfulness of the label and explanation. Users are first shown the news without access to interventions and their accuracy is measured.  We then show them the interventions and collect accuracy information again. If they do not view an intervention, no additional data is collected.  

\paragraph{Comparison with Baselines:} Table 2 shows results for all non-personalized intervention variations averaged over 3 randomized trials. For each intervention, we considered at least 604 instances of user-claim interactions (see Table \ref{table:tab4} in \ref{sec:humaneval} for full statistics of interaction instances). 

We find that without interventions, users consistently struggle to identify the true accuracy of news. All intervention types significantly improve users’ overall accuracy (up to 41.72\%). Also, all explanation-based interventions have a greater effect on accuracy than label-only interventions. 

%p{1.8cm}

\begin{figure*}[t!]
    \centering
    \begin{subfigure}[b]{0.45\textwidth}
        \centering
        \includegraphics[width=.8\linewidth]{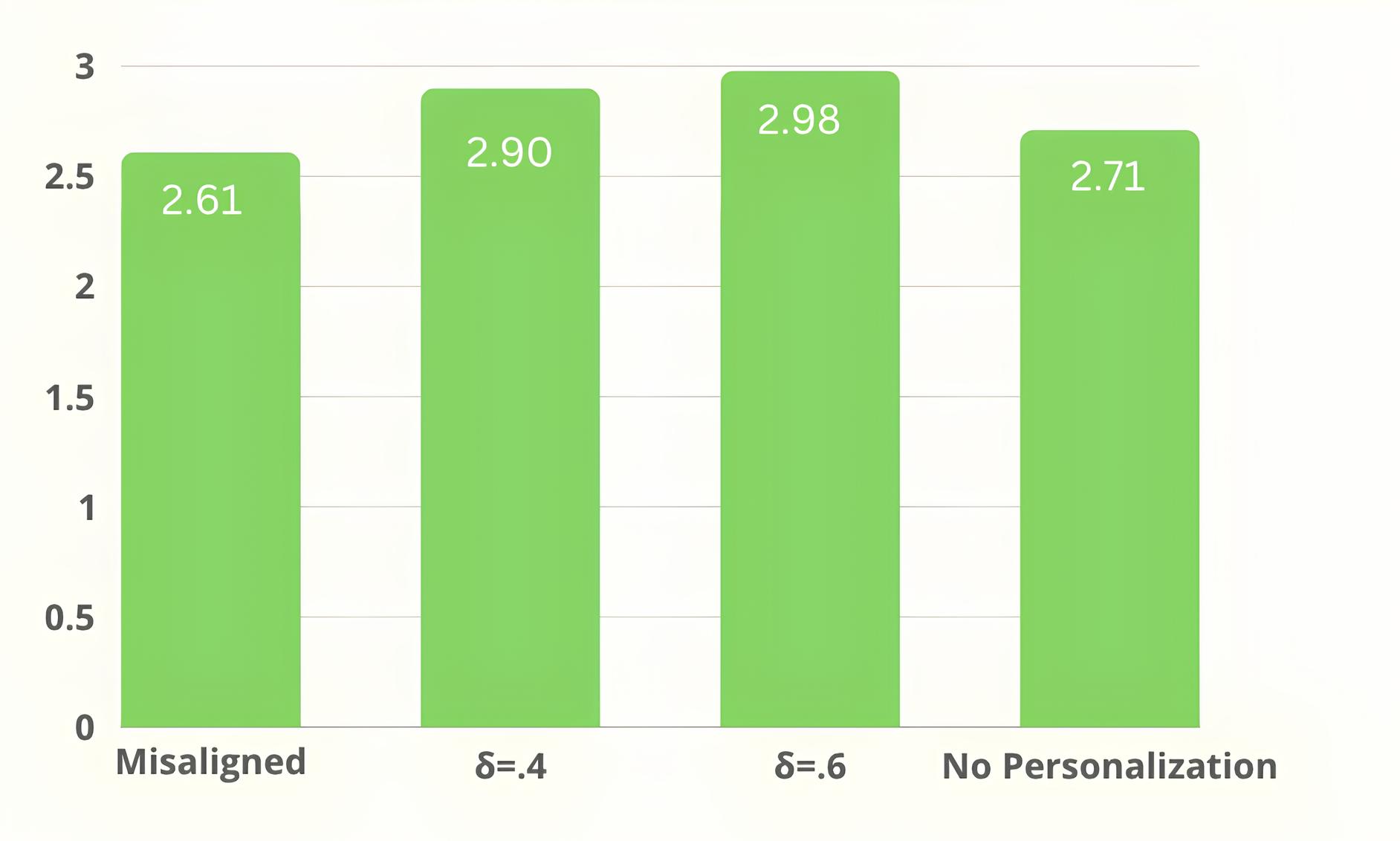}
        \caption{Mean helpfulness scores for users receiving misaligned explanations (personalization alignment score 0.2 or lower, left), aligned explanations (alignment score 0.4 or higher, center left), explanations with alignment score of 0.6 (center right), and explanations without personalization (far right).}
        \label{fig:fig5}
    \end{subfigure} \hfill
    ~ 
    \begin{subfigure}[b]{0.45\textwidth}
        \centering
        \includegraphics[width=.7\linewidth]{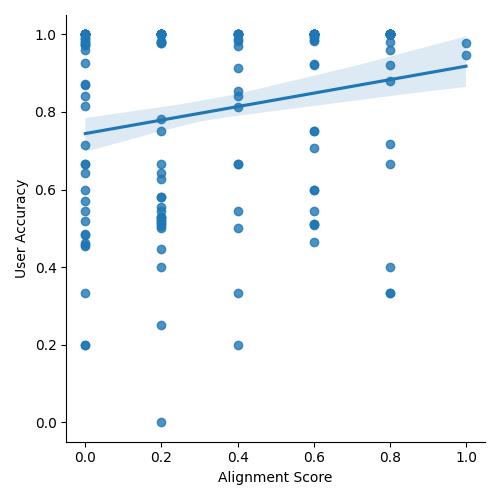}
        \caption{Linear regression analysis with 95\% confidence intervals showing explanation alignment to user attributes (x) and user accuracy (y) on a 0-1 scale. }
        \label{fig:fig6}
    \end{subfigure}
    \caption{Effects of personalization on self-reported helpfulness of explanations (left) and user accuracy (right).}
\end{figure*}

Interestingly, we find that effects on interaction behavior vary considerably across tested intervention types. In particular, interventions can actually \textit{increase} sharing of false news. One hypothesis for this may be users wanting to fact-check claims with others they trust, since we also see an increase in false content flagging for 3 out of 5 interventions. In particular, Label Only, GPT-4 and AI methodology interventions led to relatively high rates (21.38-38.17\%) of false content flagging by users.  We hypothesize that the self-contained fact-check in the GPT-4 explanation may reduce users’ interest in reaching out to trusted networks, though we do see an increase in false content sharing along with increased accuracy and flagging. 
\\
\\
Overall, three interventions seem the most effective: Reaction Frame explanations, non-personalized GPT-4 explanations, and surprisingly, a simple methodological explanation that states the veracity label was generated using an AI model, without mentioning specifics of the claim. Reaction Frame and GPT-4 explanations are comparable in terms of accuracy improvement and similar in terms of self-reported helpfulness scores, though GPT-4 explanations consistently encourage more user flagging.

\paragraph{Analysis of Healthcare Misinformation:} We further breakdown the results to assess domain-specific considerations in developing effective intervention mechanisms against medical misinformation. When we consider a balanced subset of 54 articles focused on medical real news and misinformation, we find that the prediction accuracy indicates there is no significant difference between trust in AI as a source (91.98\% acc, 95\% CI: [89.88, 94.07]) compared to a human fact-checker (92.33\% acc, 95\% CI: [90.20,94.47]). GPT-4 explanations are particularly promising interventions in the healthcare domain (97.65\% acc, 95\% CI: [96.03,99.27]), where there may be less disagreement due to partisan bias. 

\subsection{Phase II: Personalized Explanations}
\label{sec:personalized}

Our findings in the previous section indicate AI interventions can be effective at encouraging more informed, responsible behavior from users. However, this does not directly tackle risks of polarization and user bias. The next research question we seek to answer is whether GPT-4 interventions can be further improved through personalization designed to counter partisan behavior. 

\subsection{Self-reported Helpfulness Results}

\begin{figure*}
\begin{subfigure}{.5\textwidth}
  \centering
  \includegraphics[width=1\linewidth]{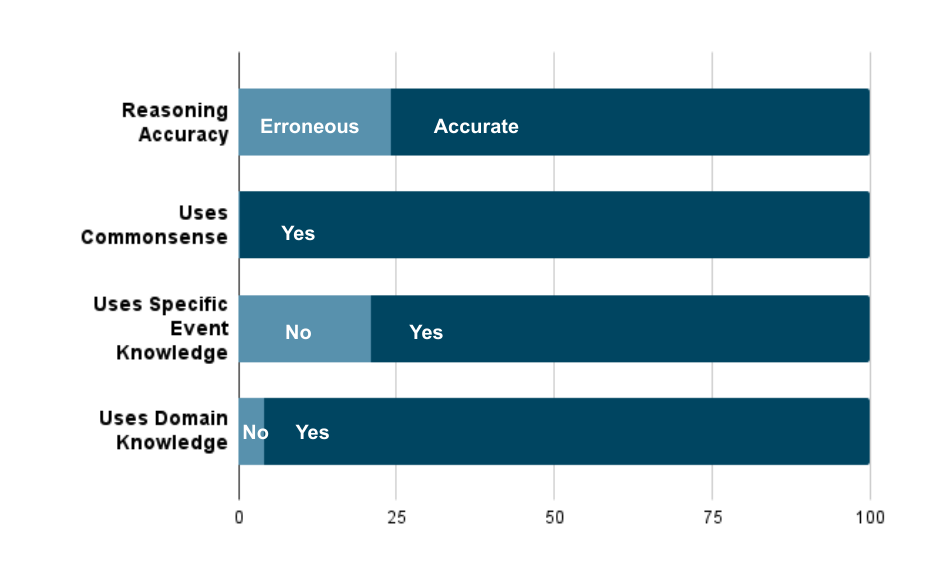}
  \caption{Breakdown of explanations by \%. }
  \label{fig:sub1-help}
\end{subfigure}%
\begin{subfigure}{.5\textwidth}
  \centering
  \includegraphics[width=.9\linewidth]{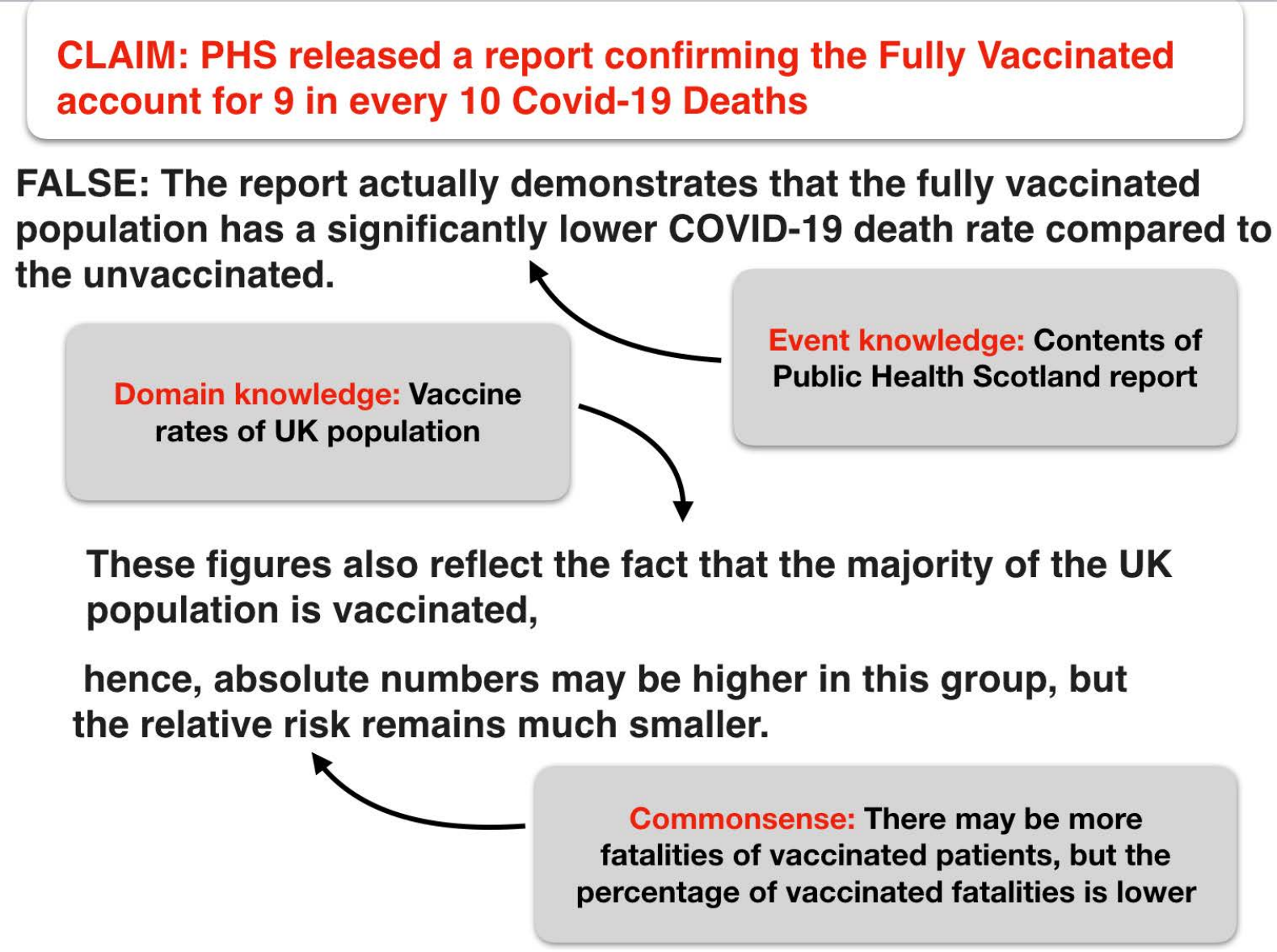}
  \caption{Example of reasoning types (commonsense, event knowledge, domain knowledge).}
  \label{fig:sub2-help}
\end{subfigure}
\caption{Analysis of all GPT-4 explanations. We provide a breakdown of explanation quality: (1) reasoning accuracy, (2) whether explanations use commonsense reasoning that should be innate to humans, (3) whether explanations require learned background knowledge from specific news events and (4) whether explanations require domain knowledge (e.g. scientific facts or knowledge of legal processes).}
\label{fig:breakdown}
\end{figure*}

We measure the effectiveness of personalization by comparing user-reported helpfulness scores for personalized explanations and scores for non-personalized GPT-4 explanations. Figure \ref{fig:fig5} shows mean helpfulness scores based on 6520 observations of GPT-4 explanations without personalization and 3000 observations with personalization. We use a 0-1 score for degree of personalization alignment, based on how many of the user's ground-truth attribute values were used to generate the explanation (e.g a score of 0.4 means 2 out of 5 attributes used to generate the explanation match the user). We consider an explanation \textit{aligned} with the user if its degree of personalization is at least = 0.4, and \textit{misaligned} otherwise. We find 49.5\% of the explanations are aligned, and the maximum alignment score is 0.6. 
\\
\\
Overall, users find explanations of veracity more helpful when they appeal to their own demographic group. As seen in Figure \ref{fig:fig5}, personalized interventions that are also aligned are given a higher mean helpfulness score ($=2.90$) than non-personalized ones ($=2.71$) ($p < .05$).\footnote{This is confirmed by both a standard t-test and Mann-Whitney U test.} Among personalized explanations, those that are sufficiently aligned with the user’s identities are also perceived to be more helpful ($=2.90$) than misaligned ones ($=2.61$). However, since self-reported helpfulness may be unreliable, we also look at accuracy.

\subsection{User Accuracy Results} To study the relationship between personalization and user accuracy, we recruit 157 participants with varying degrees of known alignment with personalized explanations for 150 claims. Figure \ref{fig:fig6} shows the linear regression analysis. For a threshold of 0.4, we find that user shown personalized explanations have an avg. accuracy of 85.89\% vs. a non-personalized user accuracy of 76.65\% (p=0.008).

\section{How Well Do LLMs Explain?}

We speculate that the strong performance described in the previous two sections is due in part to (1) the fact LLM-based interventions can convey more information than other scalable intervention types, and (2) the effectiveness of LLMs at retrieving relevant evidence from pretraining data without explicit instruction. In this section, we conduct an in-depth analysis of reasoning and form in LLM-generated explanations. We specifically address two safety risks that can arise from the way in which a language model argues for a given label. The first risk is that factually inaccurate or misleading content can arise in explanations from erroneous reasoning. The second risk is the potential for changes to linguistic properties of LLM explanations during personalization that may lead to stereotyping. 
\begin{center}
\begin{table*}
\begin{tabular}{p{.9cm} | p{4.9cm} | p{2.9cm} | p{2.6cm}| p{2.6cm}} 
 \toprule
 \small \textbf{Group} & \small \textbf{Varied attribute} & \small \textbf{Avg. length (words) $\uparrow$} & \small \textbf{Avg. readability $\uparrow$} & \small \textbf{Avg. formality $\uparrow$} \\ [0.5ex] 
 \bottomrule \toprule
 \small $g_{control}$ & \small No personalization  & \small 52.59* & \small 40.67* & \small 92.63*  \\
 \rowcolor{Gray}
 \small $g_1$ & \small Default prompt: Conservative, White, Uneducated, Male, 30-49 years old & \small 58.42 & \small 55.95 & \small 78.02  \\
 \small $g_2$ & \small Liberal  &  \small 58.45 & \small 55.99 & \small 77.84 \\
 \small $g_3$ & \small Black  &  \small 58.34 & \small 59.25* & \small 71.42* \\
 \small $g_4$ & \small Educated  &  \small 63.23* & \small 38.37* & \small 96.48* \\
 \small $g_5$ & \small Female  & \small 58.62 & \small 51.56* & \small 87.81*  \\
 \small $g_6$ & \small Age 65+ &  \small 55.98*  & \small 55.04 & \small 81.67* \\[1ex] 
 \bottomrule
\end{tabular}
\caption{ Comparison of generic GPT-4 and personalized explanations across various demographic groups using automatic metrics. Higher scores indicate greater readability or formality respectively. Statistically significant differences between $g_1$ and $g_k$ are marked by *. }
\label{table:tab3}
\end{table*}
\end{center}

\subsection{The Factuality Bottleneck}

It should be noted that users’ trust in AI is only beneficial if the model is accurate at label prediction. For the purposes of direct comparison across interventions in our study, we assume an oracle setting where the intervention label always matches the ground-truth label. We conduct a manual qualitative analysis of GPT-4 explanations generated for all true and false headlines (373 total headlines) with a Master's student who is not a co-author. We find that despite the oracle setting 24.13\% of explanations use erroneous reasoning to validate claim labels. A detailed breakdown of reasoning used in explanations is provided in Figure \ref{fig:breakdown}. We see that the model is heavily reliant on memorized event knowledge (79.09\% of explanations), which can be learned from web content like fact-checking articles and retrieved without explicit instruction from parametric knowledge. For actual deployment, steps will need to be taken to address issues of inaccurate prediction or model hallucination \cite{dziri-etal-2022-evaluating,Guan2023LanguageMH,Kalai2023CalibratedLM} due to stored parametric knowledge being out-of-date. Some potential directions include use of retrieval-augmented generation approaches \cite{Lewis2020RetrievalAugmentedGF,Nakano2021WebGPTBQ,Zhou2024CorrectingMO}. 

\subsection{Linguistic Effects of Personalization}

We compare the average length, readability and formality of personalized explanations of the ground-truth label for six different demographic groups and GPT-4 generated explanations with no personalization, denoted $g_{control}$. The first group we consider, denoted as $g_1$, has the following demographic attribute values: \textit{political affiliation = \textbf{conservative}}, \textit{race = \textbf{white}}, \textit{education = \textbf{uneducated}}, \textit{gender = \textbf{male}}, \textit{age = \textbf{30-49}}. We then consider personalized explanations for additional groups that differ from $g_1$ by exactly one attribute. Specifically, for $g_2$ \textit{political affiliation = \textbf{liberal}}, for $g_3$ \textit{race = \textbf{black}}, for $g_4$ \textit{education = \textbf{educated}}, for $g_5$ \textit{gender = \textbf{female}}, and for $g_6$ \textit{age = \textbf{65+}}. For each of these demographics, we generate personalized explanations for the ground-truth label using prompts described in §3.1. We then measure differences between explanations across groups, using length, formality prediction \cite{pavlick-tetreault-2016-empirical}, and reading difficulty based on the Flesch–Kincaid grade level metric \cite{Flesch1948ANR}. Statistical significance is assessed using a standard 2-sided t-test. 

From Table \ref{table:tab3}, we can see that lengths of explanations are relatively consistent across personalization settings. Political affiliation has the least effect across attributes, while readability and formality are significantly impacted by race, age, education and gender. In particular, specifying that the user is “educated” greatly reduces readability, indicating use of more challenging language, and increases formality by 18.46\%. Specifying that the user is “black” leads to the least formal language usage. While not inherently harmful in this setting, it does indicate potentially discriminatory assumptions held by the model that are based on the user's demographics. 

\section{Conclusion} 

In conclusion, we introduced a framework for generating, personalizing and comprehensively evaluating LLM-based misinformation interventions (MisinfoEval). Our findings show a promising direction for social media platforms and policy makers to combat misinformation by improving the presentation of content to users. With the ability of LLMs to efficiently generate explanations to support veracity judgments with user personalization, they have the potential to serve as key components in designing scalable and powerful interventions. However, it is important to highlight that their success depends on model accuracy at predicting label veracity. This requires further improvements in automated prediction tools, greater coordination with human fact-checkers, or both. However, our observations raise concern that in the future advanced LLMs like GPT-4 can be misused to create targeted “fake news” campaigns against certain groups or even individuals. The personalization ability of LLMs is a double-edged sword, and collaboration between policy makers, researchers and engineers is needed to ensure they are used for ethical and desirable intentions.

\section{Ethics Statement and Limitations}

An important aspect of (mis)information diffusion is social network interactions. It has been shown that social cues and influence of connected users contribute to spread of false content \cite{Avram2020ExposureTS,doi:10.1177/20563051231177943}. The complexity of modeling these effects renders network interaction beyond the scope of our current study. However, we encourage future work that addresses whether explanations counteract detrimental network effects. 

\paragraph{} Beyond this limitation, another concern is misuse of LLM personalization by bad actors. While our work is focused on using LLM explanations to increase online literacy and mitigate effects of misinformation, they are a potential dual-use technology. There is a need for research focused on risks of personalization being exploited to generate more persuasive misinformation and manipulate public opinion, especially given the potential use of LLMs in political campaigning \cite{Alvarez2023GenerativeAA}. As we show in §5, even in benign use cases, there are still risks of LLM deployment for content moderation like hallucinations and model demographic bias.  We hope this work facilitates development of automated personalized content moderation that bears in mind risks of stereotyping or discrimination, which has occurred in other uses of personalized LLMs \cite{wan-etal-2023-personalized}. 

\section*{Acknowledgments}

We thank David Rand for providing the data used in the study and Daniel Huttenlocher for thought-provoking discussions. We also thank colleagues at NYU, especially Julian Michael, Claudia Shi, Hannah Rose Kirk, Betty Hou, Jason Phang, and Salsabila Mahdi, for providing feedback on an early version of the study design, as well as the Dartmouth data analysis team (specifically Rong Guo) for help with summarizing experimental results. We are grateful to the anonymous ACL reviewers for their helpful comments and the MIT Generative AI Award committee for reviewing an abstract for the paper. 

%This document has been adapted
%by Steven Bethard, Ryan Cotterell and Rui Yan
%from the instructions for earlier ACL and NAACL proceedings, including those for
%ACL 2019 by Douwe Kiela and Ivan Vuli\'{c},
%NAACL 2019 by Stephanie Lukin and Alla Roskovskaya,
%ACL 2018 by Shay Cohen, Kevin Gimpel, and Wei Lu,
%NAACL 2018 by Margaret Mitchell and Stephanie Lukin,
%Bib\TeX{} suggestions for (NA)ACL 2017/2018 from Jason Eisner,
%ACL 2017 by Dan Gildea and Min-Yen Kan,
%NAACL 2017 by Margaret Mitchell,
%ACL 2012 by Maggie Li and Michael White,
%ACL 2010 by Jing-Shin Chang and Philipp Koehn,
%ACL 2008 by Johanna D. Moore, Simone Teufel, James Allan, and Sadaoki Furui,
%ACL 2005 by Hwee Tou Ng and Kemal Oflazer,
%ACL 2002 by Eugene Charniak and Dekang Lin,
%and earlier ACL and EACL formats written by several people, including
%John Chen, Henry S. Thompson and Donald Walker.
%Additional elements were taken from the formatting instructions of the \emph{International Joint Conference on Artificial Intelligence} and the \emph{Conference on Computer Vision and Pattern Recognition}.

\bibliography{acl_latex}

\clearpage

\appendix

\section{Appendix}
\label{sec:appendix}

\subsection{Explanation Generation Details}

For reproducibility, the specific GPT-4 model version we use is gpt-4-0613 with default API settings. The exact parameter count of GPT-4 is unknown. 

\subsection{Data Details}

The news claim dataset used in this work was shared by the authors of \cite{pennycook2021a} and used as intended. The dataset is entirely in English. Since the original data only contained headline images, Python-tesseract was used to extract text headlines.\footnote{\url{https://pypi.org/project/pytesseract/}}

\subsection{Other Experimental Details}

We used SciPy for statistical analysis in the paper \cite{2020SciPy-NMeth}. 

\subsection{Human Evaluation Setup}
\label{sec:humaneval}

\begin{figure*}
\centering
\begin{subfigure}{.5\textwidth}
  \centering
  \includegraphics[width=1\linewidth]{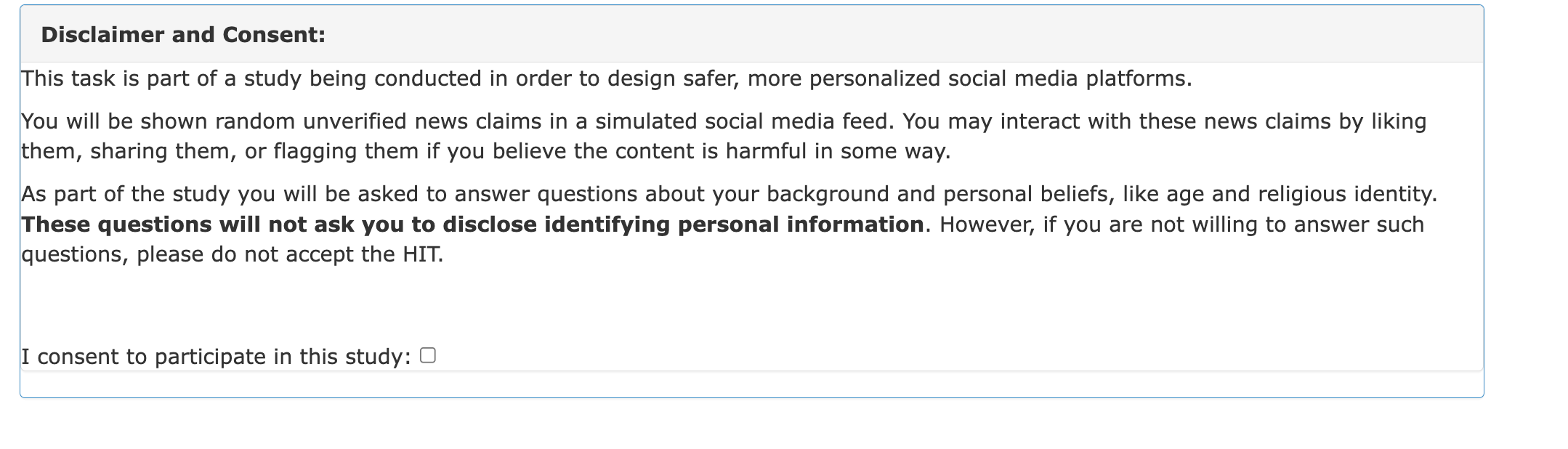}
  \label{fig:sub1-mturk}
\end{subfigure}%
\begin{subfigure}{.5\textwidth}
  \centering
  \includegraphics[width=1\linewidth]{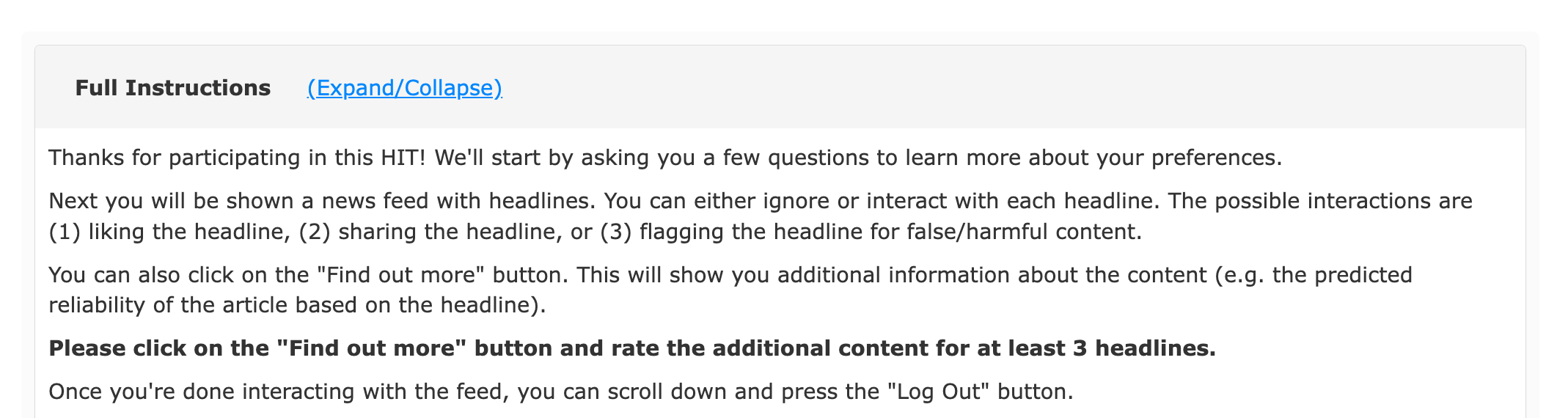}
  \label{fig:sub2-mturk}
\end{subfigure}
\caption{Full Amazon Mechanical Turk Instructions.}
\label{fig:figmturk}
\end{figure*}

We received IRB exemption approval from the MIT Committee on the Use of Humans as Experimental Subjects (COUHES) for this study. We show screenshots of full instructions to crowdworkers in Figure \ref{fig:figmturk}. All participating crowdsource workers were compensated at a rate of between \$0.40-\$1 per example depending on the study length and qualification stage, which we determined to be a fair wage given best practices for compensation of crowdsource workers, the simplicity of task and estimated time commitment \cite{10.1145/3180492}. All participants explicitly consented to take part in the task after reading a short description. The user-claim interaction instance counts for Phase I are given in Table \ref{table:tab4}. These counts enable robust statistical analysis, in particular calculation of 95\% confidence intervals shown in Table \ref{table:tab2} with a 1-3.6\% margin of error. 
\begin{center}
\begin{table}
\begin{tabular}{p{3.5cm} | p{1.5cm} | p{1.5cm} } 
 \toprule
 \textbf{Intervention Type} & \textbf{Control} & \textbf{After} \\ [0.5ex] 
 \bottomrule \toprule
  Label & 1,349  & 1,171\\
  Reaction Frame & 2,181 & 1,540 \\
  GPT-4 & 742 & 604 \\
  Methodology AI & 1,419 & 1,740\\
  Methodology Human & 2,272 & 1,655
  \\[1ex] 
 \bottomrule
\end{tabular}
\caption{Instance counts pre-intervention (Control) and post-intervention (After) for all user-claim interactions in Table \ref{table:tab2}. }
\label{table:tab4}
\end{table}
\end{center}

\subsection{Personalization Details}

\paragraph{Helpfulness Study.} For the user-reported helpfulness personalization study, we select participants based on their inferred alignment with generated personalized explanations. We generate explanations before participant selection using the following attribute sets: [conservative, uneducated, male], [moderate, white, educated, female, 30-49], [moderate, white, educated, male, 30-49], [moderate, white, educated, male, 50-64], [moderate, white, uneducated, female, 30-49], [moderate, white, uneducated, female, 50-64].\footnote{Note that due to the less diverse spread of conservative workers in our experiment, predicted when inferring attributes with Pew Research survey data, we use 3 instead of 5 attributes during personalization (political affiliation, education and gender).}

We infer user attributes by using the questionnaire in component (3) of the experiment to ask each user a list of Pew Research American Trends Panel\footnote{\url{https://www.pewresearch.org/our-methods/u-s-surveys/the-american-trends-panel/}} survey questions \cite{Santurkar2023WhoseOD} on social and political issues in the United States. We then compute the conditional probability of a person with a set of demographic attribute values giving the same answers as the user. We choose the demographic group with the highest probability. The questionnaire in component (3) of the survey also asks for the actual demographic attributes of the user for validation.\footnote{We decide to use the actual demographic values of users only for validation for several reasons. Real-world social media platforms often cannot obtain their exact values from the user or to use them in algorithms, especially for attributes like gender and race, due to privacy concerns or lack of information. This points to the need of inferring these values. Additionally, such a process allows us to perform a more fine-grained analysis on how personalization alignment affects the effectiveness of interventions.} Since they may not match the values of inferred attributes used to generate the explanation, we compute the personalization alignment score for each user, $s_{u_ie_j}$, defined as the proportion of attributes among those used to generate the personalized explanation $e_j$ that are equal to the user $u_i$’s self-reported ground-truth attributes.

\paragraph{Accuracy Study.} Participants are selected to ensure representation from left and right leaning self-reported ground-truth political ideologies, as well as unknown ideology participants. 54 participants are right-leaning and 23 are left-leaning. Explanations were personalized using the following attributes: [conservative, white, uneducated, male, 18-24] or [conservative, white, educated, male, 18-24]. 52 participants observed explanations for uneducated users and 105 observed explanations for educated users. We focus on right-learning personalization since prior research has found right-leaning users to be disproportionately targeted and involved in spread of misinformation \cite{sakketou-etal-2022-factoid, 10.1145/3555096, Pierri2022PropagandaAM}. 

\subsection{Demographics of Crowdworkers}
\label{sec:crowddemos}

We found that 2,018 workers recruited by the study answered a voluntary demographic questionnaire about age, gender, religion, politics, race and common sources for news. 52\% of workers are 25-34 years old, 30\% are 35-44 years old, 9\% are 45-54 years old, 4\% are 55-64 years old, 4\% are 18-24 years old and 1\% are over 65. 64\% of participants are male and 36\% are female.\footnote{Less than 1\% were non-binary or other.} 87\% identify as Christian, 5\% as Hindu, 3\% as Jewish, 2\% as Atheist, 2\% as Muslim and 1\% as spiritual or other. For political ideology, 29\% of participants identify as right-leaning, 26\% as moderate, 19\% as left-leaning, 18\% as very right-leaning, and 8\% as very left-leaning. For race/ethnicity, 74\% of participants identify as White, 20\% as Asian, 2\% as Black, 2\% as Hispanic, and 2\% as Native American or Pacific Islander. Most common news sources are Twitter (X), the New York Times, Breitbart, CNN, BBC News and Instagram. 

\end{document}